\documentclass[letterpaper, 10 pt, conference]{ieeeconf}  

\IEEEoverridecommandlockouts                              

\usepackage{amsmath} 
\usepackage{amssymb}  
\usepackage{multirow}
\usepackage{subcaption}
\usepackage{graphicx}
\usepackage{multicol}
\usepackage{booktabs}
\usepackage{flushend}

\title{\LARGE \bf
Tracking 6-DoF Object Motion from Events and Frames
}

\author{Haolong Li$^{1}$ and Joerg Stueckler$^{1}$
\thanks{*This work was supported by Max Planck Society and the Cyber Valley Research Fund (CyVy-RF-2019-05). 
The authors thank the International Max Planck Research School for Intelligent Systems (IMPRS-IS) for supporting Haolong Li.
}
\thanks{$^{1}$ All authors are with the Embodied Vision Group, 
        Max Planck Institute for Intelligent Systems, T\"ubingen, Germany
        {\tt\small \{haolong.li,joerg.stueckler\}@tue.mpg.de}}%
}

\begin{document}

\onecolumn
This paper has been accepted for publication in IEEE International Conference on Robotics and Automation (ICRA) 2021.

\vspace{2cm}
\textcopyright ~2021 IEEE. Personal use of this material is permitted.  Permission from IEEE must be obtained for all other uses, in any current or future media, including reprinting/republishing this material for advertising or promotional purposes, creating new collective works, for resale or redistribution to servers or lists, or reuse of any copyrighted component of this work in other works.

\twocolumn

\maketitle
\thispagestyle{empty}
\pagestyle{empty}

\begin{abstract}

Event cameras are promising devices for low latency tracking and high-dynamic range imaging.
In this paper, we propose a novel approach for 6 degree-of-freedom (6-DoF) object motion tracking that combines measurements of event and frame-based cameras.
We formulate tracking from high rate events with a probabilistic generative model of the event measurement process of the object.
On a second layer, we refine the object trajectory in slower rate image frames through direct image alignment.
We evaluate the accuracy of our approach in several object tracking scenarios with synthetic data, and
also perform experiments with real data.

\end{abstract}

\section{INTRODUCTION}

Event cameras measure intensity changes in pixels and provide an asynchronous data stream at high speed (in the microseconds). 
This brings several potential advantages over traditional frame-based cameras such as high-dynamic range, no motion blur and low latencies.
They have potential for novel robotics applications such as autonomous driving or flying robots with fast image motion or low-light settings.
Significant progress has been made in developing approaches for camera motion estimation, depth reconstruction, and high dynamic image reconstruction with event-based cameras~\cite{gallego2020_evsurvey}.
The measurement principle can also provide novel opportunities for 6-DoF object tracking which is still largely unexplored by the research community. 

In this paper, we propose a novel approach for 6-DoF object tracking with event cameras.
Object pose tracking is an important perception capability in many robotics applications in dynamic scenes, for instance, for dynamic obstacle avoidance or robotic manipulation.
In our tracking approach, we leverage a combination of event and frame-based camera measurements which can be obtained, for example, from event-based cameras such as the iniVation DAVIS camera that also provides intensity frame measurements, or by pairing an event camera with a frame-based camera.
Our approach tracks objects with known shape which can be given by a mesh.
We derive a probabilistic measurement model for the object's shape under rigid-body motion and track the motion of the object from the event stream using probabilistic inference.
On a second optimization layer, we refine the tracked object poses using probabilistic direct image alignment in a window of frames captured by the camera at lower rates.
We evaluate our approach on synthetic scenes with moving objects in indoor and outdoor scenarios and analyze the accuracy of our approach for 6-DoF pose tracking.
We also demonstrate our approach in experiments with real data.


\begin{figure*}[tb]
	\centering
	\includegraphics[width=0.49\linewidth]{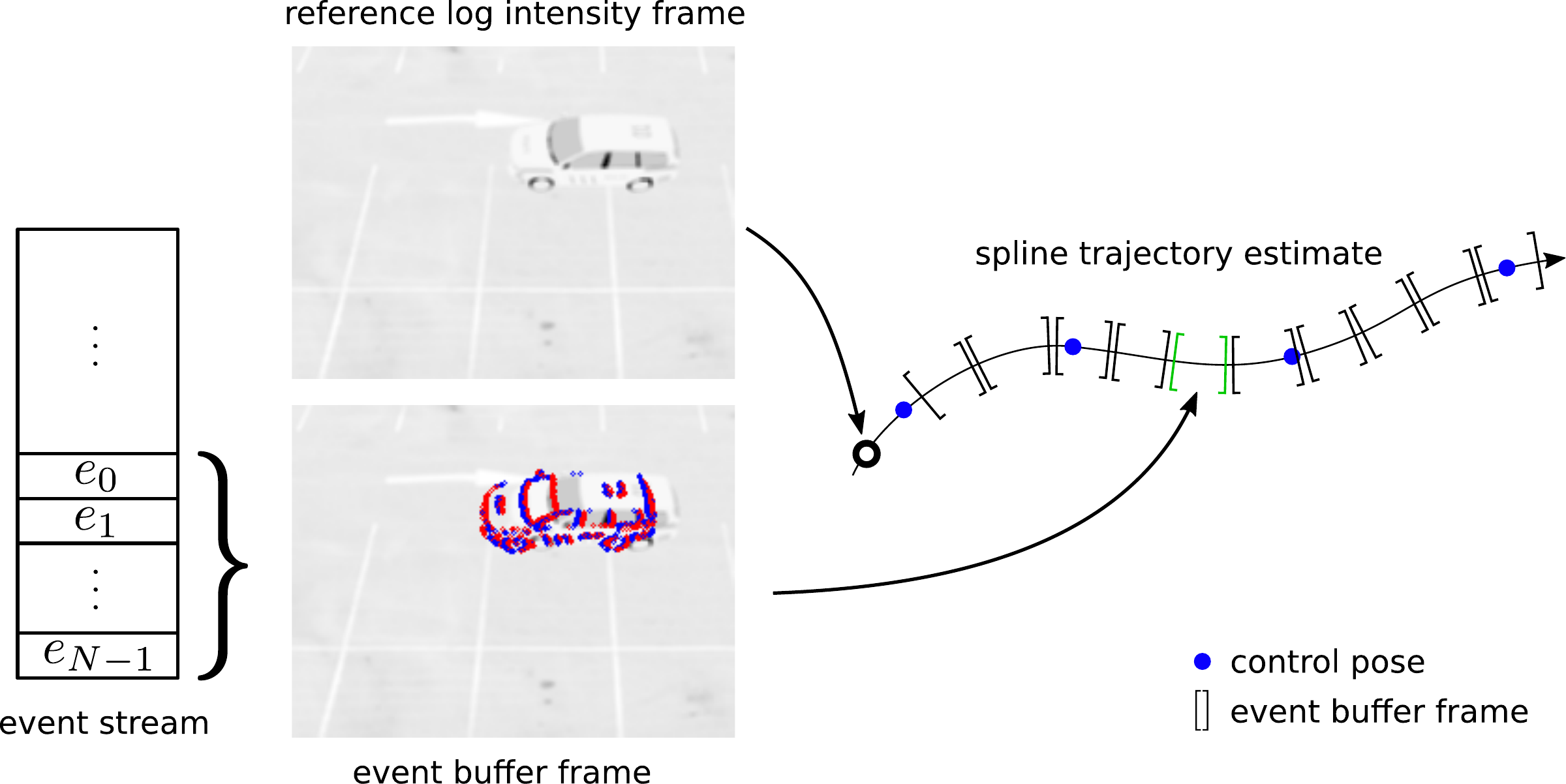}\hfill		
	\includegraphics[width=0.49\linewidth]{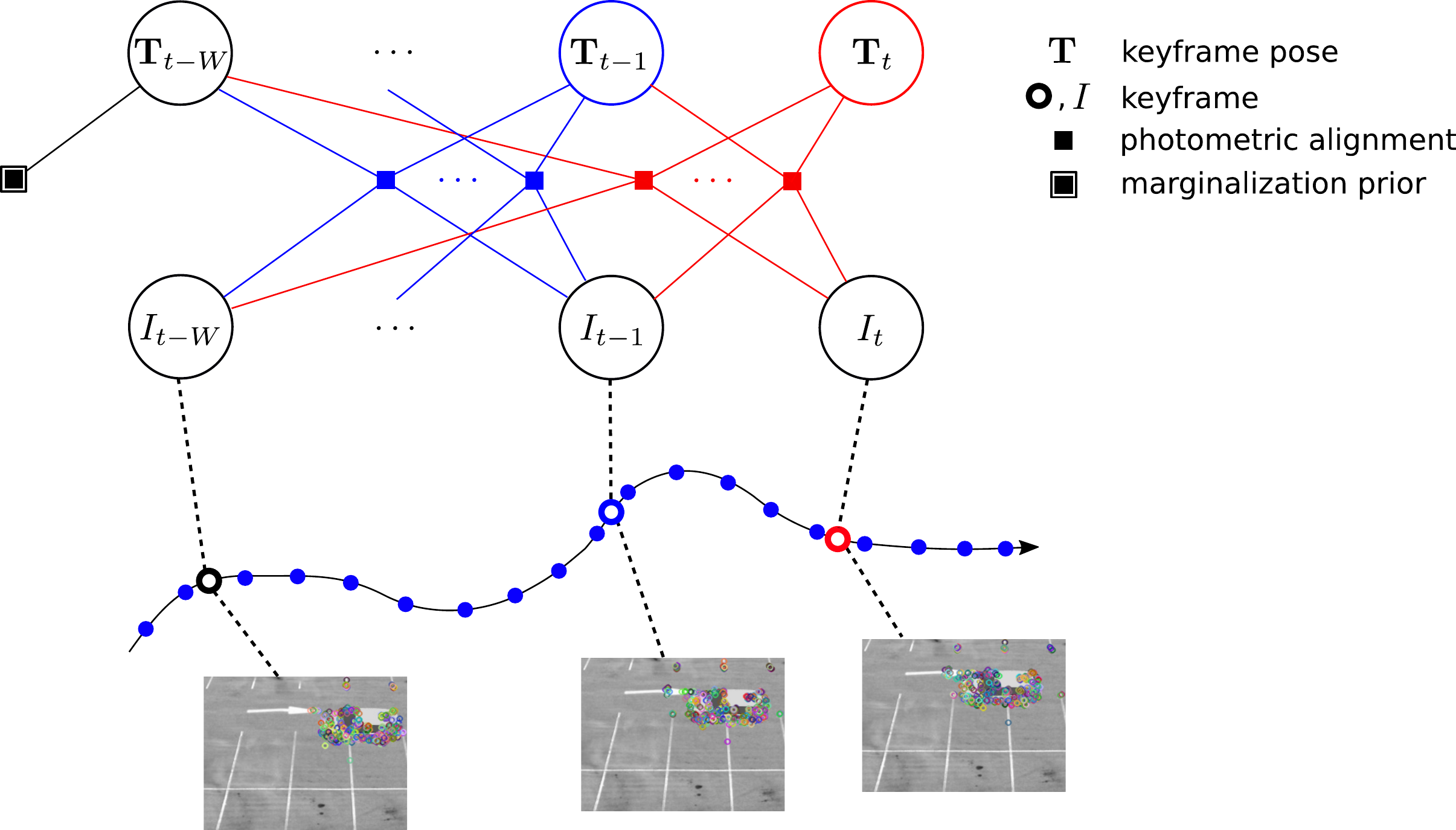}	
	\caption{Left: We track the motion of the object from the stream of events. We parametrize the motion using a cubic B-spline. The control poses are optimized based on a probabilistic generative event measurement model. The model predicts intensity changes using the object velocity and the intensity gradient in a reference frame. For computational efficiency, we accumulate $N$ consecutive events $e_i$ in event buffer frames. Right: We refine the object pose estimates of keyframes extracted from the images of a frame-based camera. We align the images photometrically at keypoints based on the known object shape and the estimated object poses. The latest keyframe and its estimated pose serves as reference for event-based tracking.}
	\label{fig:overview}
\end{figure*}

\section{Related Work}

Event-based cameras have several potential advantages over frame-based cameras for tracking:
They provide high temporal resolution in the microseconds, high dynamic range, and low power consumption~\cite{gallego2020_evsurvey}.
Hence, methods for processing and interpreting event-streams have become a popular subject in computer vision and robotics research.

While significant research has been devoted to localizing event cameras in a given 3D map or simultaneous localization and mapping (e.g.~\cite{kim2016_slamtracking,rebecq2017_evo}), only few works consider the problem of object tracking.
Vasco et al.~\cite{vasco2017_evmodet} detect independent moving objects by tracking corners detected in event images integrated over short time windows.
Mitrokhin et al.~\cite{mitrokhin2018_evmodt} estimate a warp field for the events in a time window using a 4-parameter global motion model. 
Using a time image representation, they segment events on moving objects which do not comply with the motion in the estimated background motion model.
\cite{mitrokhin2019_evimo} train a neural network which predicts depth images, motion estimates and motion segmentation from integrated event images and their time image in a time window.
Different to our tracking approach, these methods estimate the 2D motion of the object.

Very recently, Dubeau et al.~\cite{dubeau2020_rgbde} extend a deep learning-based object tracker for RGB-D cameras by also incorporating events from an event camera for high-speed object tracking.
In contrast to this learning-based approach, we model the measurement process in a probabilistic generative way and derive an optimization-based method which combines events and frames.
This way, our approach is not limited by the variation of objects and scenes which are seen by a deep neural network during training.

Some approaches for localization and mapping with event cameras have several similarities with our method.
Kim et al.~\cite{kim2016_slamtracking} estimate 6-DoF camera motion, log intensity gradient and inverse depth using three decoupled probabilistic filters in real-time.
Using the log intensity gradient, high-dynamic range log intensity frames can be recovered through convex optimization.
The methods uses a probabilistic measurement model similar to ours which is derived from the event-generation process.
For camera motion tracking, the method assumes depth and log intensity of the scene given from the other filtering steps and uses the estimates to raycast intensity changes in the camera motion estimate.
These expected measurements are compared with the intensity changes detected by the events to filter the camera pose.
In our approach, we also recover camera motion by coupling intensity gradient with optical flow and use a probabilistic filter to estimate the camera motion.
Bryner et al.~\cite{bryner2019_eventbasedtracking} propose a method for tracking camera motion with regard to a static background which has been captured using an RGB-D camera.
They also apply a variant of the generative event measurement model used in~\cite{kim2016_slamtracking} and our approach.
Different to their method, we track and segment objects in the event stream.
Moreover, we combine event and frame-based tracking in a two-layer optimization approach. 

\section{Method}

We propose a novel approach which tracks the 3D motion of objects in a combined way from measurements of event and frame-based cameras.
We formulate tracking as optimization using a probabilistic measurement model of the event generation process.
In a second optimization thread, we refine the object poses in the image frames using direct image alignment.


Fig.~\ref{fig:overview} illustrates the key steps of our method.
The camera provides an asynchronous stream of event measurements together with a stream of intensity frames at a regular slower rate. 
We use both types of data for tracking the 6-DoF object motion.
On a fast optimization layer, we fit an $SE(3)$ trajectory spline to the events measured on the object.
We formulate event-based tracking as a probabilistic inference problem based on a generative model of the event measurement process.
Inference is achieved by non-linear least squares to optimize the spline control poses.

A second slower optimization layer refines the object poses in the intensity frames measured by the camera. 
To this end, we formulate the optimization again as a probabilistic inference, whereas now we measure the photometric alignment of the images using the object pose and shape in the frames.
This optimization layer provides the reference object poses and intensity images for the event-based tracking layer.
The event-based tracker on the other hand provides the frame-based optimization layer with initial poses of the objects in the frames and achieves higher frame-rate high-speed tracking.

\subsection{Generative Event Model}

We use two kinds of measurements models to explain the stochastic generative process underlying the events.

\subsubsection{Intensity Change Measurement Model}
Let $\mathbf{u}(t)$ be the image projection of 3D points in the scene.
The image projection is changing over time due to the underlying motion of the camera or the object.
Assuming the brightness of a 3D point observation in the image remains constant, the measured intensity at a corresponding pixel location at different time steps $t, t'$ stays equal, i.e.
	$L( \mathbf{u}(t), t ) = L( \mathbf{u}(t'), t' )$.	
Linearizing the log intensity image $L$ for the time using a first order Taylor approximation yields
\begin{equation}
	L( \mathbf{u}(t+\delta t), t+\delta t ) \approx L( \mathbf{u}(t), t ) + \frac{\partial L}{\partial \mathbf{u}} \frac{\partial \mathbf{u}}{\partial t} \delta t + \frac{\partial L}{\partial t} \delta t.
\end{equation}
Hence, we obtain the optical flow constraint
\begin{equation}
	\frac{\partial L}{\partial t}( \mathbf{u}(t), t ) + \nabla L( \mathbf{u}(t), t ) \dot{\mathbf{u}}(t) = 0
\end{equation}
which relates the intensity change over time with the spatial intensity change and the optical flow in the image.

Events measure intensity changes at pixels $\mathbf{u}$.
By approximating the intensity change at a pixel
\begin{equation}
	\Delta L( \mathbf{u}, t ) = L( \mathbf{u}, t ) - L( \mathbf{u}, t - \Delta t ) \approx \frac{\partial L}{\partial t}( \mathbf{u}(t), t ) \Delta t,
\end{equation}
we can rewrite the change using the optical flow constraint as
	$\Delta L( \mathbf{u}, t ) \approx - \nabla L(\mathbf{u}) \dot{\mathbf{u}} \Delta t$.
This relation explains the intensity changes generating events by the component of the flow $\dot{\mathbf{u}}$ along the image gradient $\nabla L(\mathbf{u})$ in the time interval $\Delta t$~\cite{kim2016_slamtracking,bryner2019_eventbasedtracking}.
The intensity change measurement model is
\begin{equation}
\label{eq:intchg_meas_model}
	y = C = -\nabla L(\mathbf{u}) \dot{\mathbf{u}} \Delta t + \delta,
\end{equation}
where $\delta \sim \mathcal{N}( 0, \mathbf{Q} )$ models measurement noise, and $C$ is the log intensity change for the event. 
The optical flow $\dot{\mathbf{u}}$ is determined by the camera velocity and the depth $d:=d(\mathbf{u})$ at the pixel $\mathbf{u}:=(x,y)^\top$~\cite{corke2013_rvc}
\begin{multline}\arraycolsep=6pt
	B(\mathbf{u}) := \\ \left( \begin{array}{cccccc}
		-1/d & 0 & x / d & xy & -1-x^2 & y\\
		0 & -1/d & y/d & 1+y^2 & -xy & x\\
	\end{array} \right).
\end{multline}

Since the event stream does not provide the image gradient directly, we determine the image gradient in a keyframe. The image gradient can be obtained from an intensity frame as measured for instance by a frame-based camera.
In our experiments, we use the intensity frames measured concurrently with the events by a (simulated) DAVIS240C camera.
The image gradient is determined by reprojecting the pixel with its raycasted depth on the object in the keyframe.

\subsection{Event-based Object Tracking}

We track the motion of the object using the asynchronous event stream.
We choose to increase the computational efficiency and average over noisy individual event readings by buffering multiple events at the pixels over short time windows.
We use a continuous SE3 spline to represent the object trajectory.
The spline provides pose estimates at arbitrary continuous times with a small set of control poses which we leverage for the asynchronous event buffer frames.
The spline also inherently regularizes the estimated trajectory to be smooth.
We formulate a probabilistic optimization objective to fit the spline trajectory to the event stream.

%
%

\subsubsection{Spline Trajectory Representation}

We represent the time-continuous pose $\mathbf{T}(t) \in SE(3)$ at time $t \in \mathbb{R}^3$ with a cubic B-spline in cumulative form~\cite{patronperez2015_splinetraj},
\begin{equation}
	\mathbf{T}(t) = \exp\left( \log\left( \mathbf{T}_0 \right) \right) \prod_{k=1}^{K-1} \exp\left( \log\left( \mathbf{T}_{k-1}^{-1} \mathbf{T}_k \right) B_k(t) \right)
\end{equation}
where $\mathbf{T}_k$ are control poses at knot times $t_k$ and $B_k(t)$ are cumulative basis functions.

\subsubsection{Event-based Trajectory Optimization}

We optimize the spline segment of the four most recent control poses from all events in the segment.
To increase computational efficiency and cancel noise due to the image discretization, we accumulate event buffer frames from the events in short time windows.

\paragraph{Event Buffer Frames}
We accumulate the intensity changes of $N$ consecutive events $e_0, \ldots, e_{N-1}$ in event buffer frames $F$.
Events $e_i = ( x_i, y_i, \rho_i ) \in \mathcal{E}$ are generated at pixel $(x_i,y_i)$ and provide the sign $\rho_i$ of the log intensity change with magnitude $C$.
Since we assume consecutive events, it holds $t_0 \leq \ldots \leq t_{N-1}$.
Pixels $F( x, y )$ of the event buffer frame accumulate intensity changes 
\begin{equation}
	F( x, y ) = \sum_{\{e_i \in \mathcal{E} : (x_i,y_i) = (x,y)\}} \rho_i C.
\end{equation}
We refer to the time $t_0$ of the first event in the event frame buffer as its start time $t_F := t_0$ and $\Delta t_F := t_{N-1} - t_0$ is the time span of the events in the buffer.
If $N$ events are accumulated, a new buffer is started.

\paragraph{Trajectory Optimization}

We optimize the control poses of the most recent spline segment when the segment is filled with event buffer frames, i.e. when the start time of the newest event buffer frame becomes later than the knot time $t_{K-2}$ of the second to last control pose $\mathbf{T}_{K-2}$.
After optimization, we include a new control pose at a fixed time interval.
The new segment shifted by a control pose is again optimized as soon as it is filled.

We optimize the objective function 
\begin{equation}
	E( \mathbf{T}_{K-4}, \ldots, \mathbf{T}_{K-1} ) =
	E_{\mathit{data}} + \lambda_{\mathit{reg}} E_{\mathit{reg}}
\end{equation}
which consists of a data term $E_{\mathit{data}}$ and regularization term $E_{\mathit{reg}}$ with weighting factor $\lambda_{\mathit{reg}}$, which we derive from the maximum a posterior estimate of the control poses given the event buffer frame measurements and a regularizing prior on the object acceleration.

The data log likelihood of the event buffer frames within the spline segment are determined by the probabilistic intensity change measurement in eq.~\eqref{eq:intchg_meas_model},
\begin{multline}
E_{\mathit{data}} = 
\sum_{F \in \mathcal{F}} \sum_{\mathbf{u} \in \Omega} \frac{w_c^2}{w_c^2 + \left\| \nabla L(\mathbf{u}) \right\|_2^2} \\
\left\lVert \frac{F( \mathbf{u} )}{ \sqrt{ \sum_{\mathbf{u'} \in \Omega} \left(F( \mathbf{u}' )\right)^2 } }  + \frac{\nabla L(\mathbf{u}) \dot{\mathbf{u}}(t_F) \Delta t_F}{\sqrt{ \sum_{\mathbf{u'} \in \Omega} \left( \nabla L(\mathbf{u}') \dot{\mathbf{u}'}( t_F ) \Delta t_F \right)^2 } }   \right\rVert ^2 
\end{multline}
where $\mathcal{F}$ is the set of event buffer frames in the spline segment, $\Omega$ is the set of pixel coordinates in the image.
We follow the approach of~\cite{bryner2019_eventbasedtracking} and normalize the intensity changes due to the unknown log intensity threshold $C$ of the camera in practice.
Similar to~\cite{engel2018_dso}, we downweight pixels with high gradient where $w_c$ determines the strength of this factor.
For evaluating the expected intensity changes $\nabla L(\mathbf{u})$, we use the last keyframe $I_{\mathit{KF}}$ and its pose from the frame-based photometric optimization layer,
	$\nabla_{\mathbf{u}} L(\mathbf{u}) \approx \nabla_{\mathbf{u}} L_{KF}( \tau( \mathbf{u}, \mathbf{T}(t_F) ) )$,
where $L_{\mathit{KF}} = \log( I_{\mathit{KF}} )$.
The event pixel location is projected to the keyframe as
\begin{equation}
	\tau( \mathbf{u}, \mathbf{T}(t_F) ) = \pi\left( \mathbf{R}_F^{KF} \pi^{-1}\left( \mathbf{u}, d(\mathbf{u}, \mathbf{T}(t_F), \mathbf{\Phi} ) \right) + \mathbf{t}_F^{KF} \right),
\end{equation}
where $\pi$ and $\pi^{-1}$ project 3D coordinates to image pixels and vice versa using the known camera calibration, the latter requiring the depth at the pixel.
We raycast the depth $d(\mathbf{u}, \mathbf{T}_F^{KF}, \mathbf{\Phi} )$ on the object shape in the given pose which can be differentiated for the pose as in~\cite{wang2020_directshape}. 
The shape is represented by the signed distance function $\mathbf{\Phi}$.
3D points in camera coordinates are transformed between the frames using the rotation $\mathbf{R}_F^{KF} \in SO(3)$ and translation $\mathbf{t}_F^{KF} \in \mathbb{R}^3$ of the $SE(3)$ transform $\mathbf{T}_F^{KF} = \mathbf{T}_{KF} \mathbf{T}(t_F)^{-1}$ determined from the object pose in the keyframe $T_{KF}$ and the spline.

The acceleration prior
$	E_{\mathit{reg}} = \sum_{F \in \mathcal{F}} \left\lVert \ddot{\mathbf{T}}(t_F) \right\rVert_2^2
$
favors constant velocity trajectories and is evaluated at the start times of the event buffer frames.

We assume an initial guess of the object pose known with which we initialize the first four control poses. 
To this end, we use ground truth in our current implementation. 
Note that accurate frame-based object detection and pose estimation might also be used.

\subsection{Keyframe-based Photometric Trajectory Optimization}

A second layer optimizes for the poses of the keyframes which are used as reference for the event-based tracking.
Based on the known object shape, we estimate object poses in the keyframes which best align the keyframe images photometrically.
The tracked object pose from the event-based tracking layer is used to initialize the object pose in new keyframes.
New keyframes are selected from the intensity frames based on thresholds on rotation and translation. 
We optimize the object trajectory in a window of recent keyframes and marginalize old keyframes that shift outside the window.

\subsubsection{Photometric Alignment}

The object pose is determined in the keyframes by photometric alignment between pairs of keyframes in the optimization window.
From each keyframe, we extract ORB keypoints~\cite{rublee2011_orb} and limit computation to only those points which project within a soft silhouette mask of the object given its current pose estimate. 
Photometric residuals are computed at the remaining keypoints on the object to optimize for the object pose. 

\paragraph{Silhouette Projection}
We adapt the projection function proposed by~\cite{prisacariu2012_segposerecon,dame2013_denserecon}
\begin{equation}
	\sigma( \mathbf{\Phi}, \mathbf{u} ) = 1 - \prod_{\mathbf{p} \sim \mathcal{R}(\mathbf{u})} \frac{ \exp\left( \mathbf{\Phi}( \mathbf{p} ) \zeta \right) }{ \exp\left( \mathbf{\Phi}( \mathbf{p} ) \zeta \right) + 1}
\end{equation}
which projects the object shape, represented by the signed distance function $\mathbf{\Phi}$, into the image to find a smooth silhouette.
The projection samples points $\mathbf{p}$ on the ray $\mathcal{R}(\mathbf{u})$ through the pixel $\mathbf{u}$ and evaluates a smooth indicator function for each point being inside or outside the object based on the signed distance function.
We threshold at a specific value ($0.95$ in our experiments) and discard all ORB keypoints that project outside the object.

\paragraph{Photometric Residuals}
We formulate a photometric measurement model at the keypoints in each keyframe $I_i$ towards other keyframes $I_j$
\begin{equation}
\label{eq:photometricerror}
	z = I_i( \mathbf{u} ) = I_j( \omega( \mathbf{u}, \mathbf{T}_i, \mathbf{T}_j ) ) + \epsilon
\end{equation}
where $\epsilon \sim \mathcal{N}( 0, \sigma_I^2 )$ is Gaussian noise and $\omega$ reprojects pixels $\mathbf{u}$ on the object from image $I$ to $I'$ based on the object shape and the poses $\mathbf{T}_i, \mathbf{T}_i$ of the keyframes.
We reproject pixels through the warping function
\begin{equation*}
	\omega( \mathbf{u}, \mathbf{T}_i, \mathbf{T}_j ) = \pi\left( \mathbf{R}_i^j \pi^{-1}( \mathbf{u}, d(\mathbf{u}, \mathbf{T}_i, \mathbf{\Phi} ) ) + \mathbf{t}_i^j \right).
\end{equation*}
Here, we use the $SE(3)$ transform $\mathbf{T}_i^j = \mathbf{T}_i \mathbf{T}_j^{-1}$ determined from the object poses in the frames to transform pixels.
Keypoints are discarded which do not hit the object shape during raycasting for depth estimation.

\subsubsection{Windowed Optimization}

We formulate the optimization for the object poses in the keyframes as maximum a posteriori estimation using the probabilistic photometric measurement model
\begin{equation}
	E\left( \{ \mathbf{T}_i \}_{i=0}^{M-1} \right) = \sum_{i=0}^{M-1} \sum_{j < i}\rho\left( I_i( \mathbf{u} ) - I_j( \omega( \mathbf{u}, \mathbf{T}_i, \mathbf{T}_j ) ) \right)
\end{equation}
where we determine photometric residuals from a keyframe to older keyframes, and $\rho$ is the robust Cauchy norm.

For bounding the run-time complexity, the keyframe poses are optimized in a sliding window of a fixed number $W=7$ of keyframes similar to~\cite{leutenegger2015_okvis}.
Information about old keyframes that drop out of the sliding window is marginalized in our probabilistic formulation.
The optimization is triggered only if a new keyframe is inserted. 
Once the optimization converged, the last keyframe is passed to the event-based tracking layer as the new reference frame.

\subsubsection{Keyframe Selection}

We select the latest current frame as a new keyframe, if the translation or rotational distance travelled by the object according to the event-based tracking layer are larger than some thresholds.
The pose of a new keyframe is initialized with the current tracked pose of the object by the event-based tracking.

\begin{figure*}[tb]
	\centering
	\begin{subfigure}{.15\textwidth}
		\includegraphics[width=\linewidth]{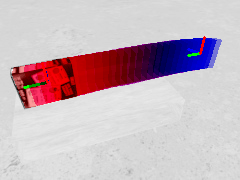}\\\vspace{-1.6ex}
		\includegraphics[width=\linewidth]{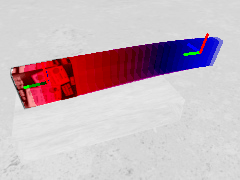}
	\end{subfigure}
	\begin{subfigure}{.15\textwidth}
	\includegraphics[width=\linewidth]{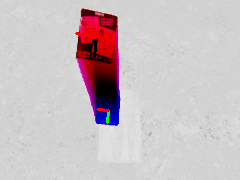}\\\vspace{-1.6ex}
		\includegraphics[width=\linewidth]{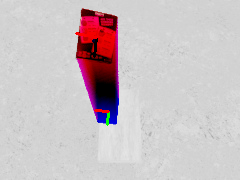}
	\end{subfigure}
	\hspace{2ex}
	\begin{subfigure}{.15\textwidth}
		\includegraphics[width=\linewidth]{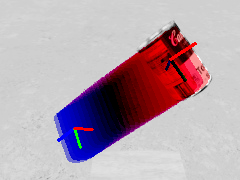}\\\vspace{-1.6ex}
		\includegraphics[width=\linewidth]{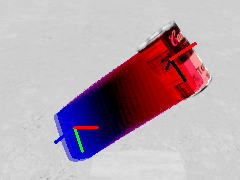}
	\end{subfigure}
	\hspace{2ex}
	\begin{subfigure}{.15\textwidth}
		\includegraphics[width=\linewidth]{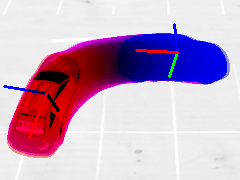}\\\vspace{-1.6ex}
		\includegraphics[width=\linewidth]{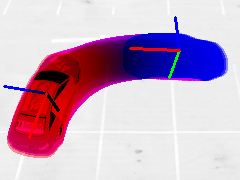}
	\end{subfigure}
	\begin{subfigure}{.15\textwidth}
		\includegraphics[width=\linewidth]{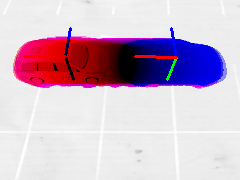}\\\vspace{-1.6ex}
		\includegraphics[width=\linewidth]{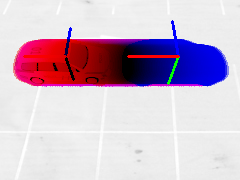}
	\end{subfigure}

	\caption{Estimated (top) vs. ground-truth trajectory (bottom) as image overlays for the YCB box object (left two columns: sliding, falling), the YCB can object (middle column: dice) and the two car sequences (right two columns). Time is visualized from red to blue for start to end. Our approach well recovers the ground-truth motion of the objects. 
	}
	\label{fig:trajectoryoverlays}
\end{figure*}

\begin{figure}[tb]
	\centering
	\begin{subfigure}{.99\linewidth}
	\centering
		\includegraphics[width=0.3\linewidth]{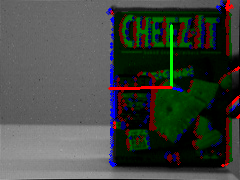}
		\includegraphics[width=0.3\linewidth]{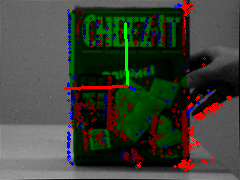}
		\includegraphics[width=0.3\linewidth]{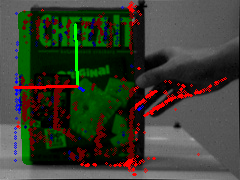}
	\end{subfigure}\\\vspace{1ex}
	\begin{subfigure}{.99\linewidth}
	\centering
		\includegraphics[width=0.3\linewidth]{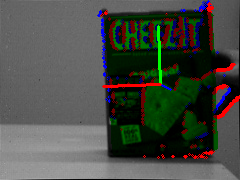}
		\includegraphics[width=0.3\linewidth]{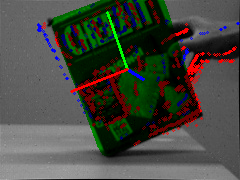}
		\includegraphics[width=0.3\linewidth]{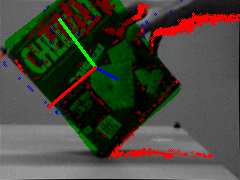}
	\end{subfigure}\\\vspace{1ex}
	\begin{subfigure}{.99\linewidth}
	\centering
		\includegraphics[width=0.3\linewidth]{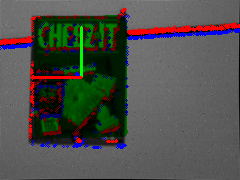}
		\includegraphics[width=0.3\linewidth]{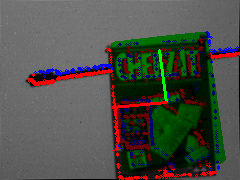}
		\includegraphics[width=0.3\linewidth]{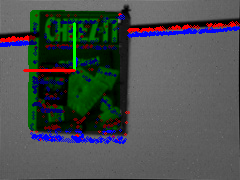}
	\end{subfigure}
	\caption{Results on real data sequences with translational (top), rotational (middle) and circular motion (bottom).  Left: start, right: end frame.
	Green shaded/axes: estimated object pose; red/blue points: positive/negative events.
	}
	\label{fig:realdataresults}
\end{figure}

\section{Experiments}

We evaluate our approach for tracking accuracy on synthetic datasets where ground truth poses are available.
We use the standard absolute trajectory (ATE) and relative pose error measures~\cite{sturm2012_iros} which evaluate the global alignment of the trajectory and drift, respectively. 
We evaluate RPE for $10,20,\ldots,50$\% sequence lengths of the full trajectory length in m. 
We obtain a single RPE measure by averaging results over all sequence lengths. 
Qualitative results can also be found in the supplementary video.


Our synthetic sequences are generated with an event camera simulator~\cite{rebecq2018_cvsimulator} based on Blender. 
We generate sequences for 2 objects from the YCB set~\cite{calli2015_ycb} and 2 cars from the ShapeNet dataset~\cite{chang2015_shapenet}.
For the YCB objects, we have 3 sequences each with falling, sliding and dice motion.
The cars move either straight, turn left or turn right, whereas we use 3 different speed settings per sequence type.

\subsection{Setup}

The time span between knots in the spline is set to 15\,ms, and we set $\lambda_{\mathit{reg}} = 0.1$. 
The frame rate of the camera is 30\,Hz.
For the YCB objects we accumulate $N=1500$ in each event buffer frame. 
We generate new keyframes after the object travelled a translational distance of 0.1\,m and a rotational distance of 5\,deg. The contrast weight factor $w_c$ is $0.005$. 
For the car objects we create the motions with 3 different speed levels. In the first level the average linear speed is about 1.25\,m/s. The second speed level is about 2.5\,m/s, and the third speed level is about 5\,m/s. The pose is initialized without noises.
The translational distance of keyframes is 0.5\,m and the rotational distance is 10\,deg. We set the number of events per event buffer frame to $N=4000$. For the textureless car the contrast weight factor $w_c$ is 0.3 and for the textured one it is set to 0.1.

\subsection{Results}

\subsubsection{Synthetic YCB object sequences}

Table~\ref{tab:accuracy_ycb} shows RPE and ATE results for the various sequences with the box and can YCB objects.
The object maximum sizes are 0.3\,m and 0.45\,m. The camera is positioned above the objects.
The average distance of the objects from the camera is 1.6\,m for the sequences.
We observe that position and rotation of the objects are tracked by our approach at good accuracy on these sequences.
Fig.~\ref{fig:trajectoryoverlays} depicts trajectory overlays of the estimates by our approach and the ground-truth.
It can be seen that our approach well recovers the object motion.


\begin{table*}[tb]
	\caption{Accuracy of trajectory estimates in relative pose (RPE) and absolute trajectory error (ATE) on YCB sequences. Our tracking approach recovers the object motion at good accuracy.}
	\label{tab:accuracy_ycb}
	\begin{center}
		\begin{tabular}{cccccccccc}
			\toprule
			 & \multicolumn{3}{c}{transl. RMSE RPE in\,m} &
			\multicolumn{3}{c}{rot. RMSE RPE in\,deg} &
			\multicolumn{3}{c}{transl. RMSE ATE in\,m}\\
			\cmidrule(lr){2-4} \cmidrule(lr){5-7} \cmidrule(lr){8-10}
			object & falling & sliding & dice  & falling & sliding & dice & falling & sliding & dice\\
			\midrule
			box & 0.086 & 0.056 & 0.024 & 5.196 & 5.686 & 3.840 & 0.064 & 0.028 &  0.012\\
			can & 0.038 & 0.107 & 0.012 & 5.640  & 6.110 & 1.423 & 0.029 & 0.119 & 0.031 \\
			\bottomrule
		\end{tabular}
	\end{center}
\end{table*}

\begin{table*}[tb]
	\caption{Accuracy of trajectory estimates in relative pose (RPE) and absolute trajectory error (ATE) on car sequences}
	\label{tab:accuracy_cars}
	\begin{center}
		\begin{tabular}{ccccccccccc}
			\toprule
			 & &\multicolumn{3}{c}{transl. RMSE RPE in\,m} &
			\multicolumn{3}{c}{rot. RMSE RPE in\,deg} &
			\multicolumn{3}{c}{transl. RMSE ATE in\,m}\\
			\cmidrule(lr){3-5} \cmidrule(lr){6-8} \cmidrule(lr){9-11} 
			object & speed & left turn & right turn & straight & left turn & right turn & straight & left turn & right turn & straight \\
			\midrule
			textured & 1x & 0.189 & 0.283 & 0.250 & 2.764 & 3.306 & 2.958 & 0.097 & 0.171 & 0.149\\
			textured & 2x & 0.409 & 0.355 & 0.206 & 5.612 & 2.991 & 2.876 & 0.114 & 0.217 & 0.109\\
			textured & 4x & 0.181 & 0.406 & 0.453 & 2.957 & 3.033 & 3.478 & 0.080 & 0.175 & 0.138\\
			\midrule
			textureless & 1x & 0.811 & 0.450 & 0.528 & 10.816  & 6.168 & 6.245 & 0.471 & 0.224 & 0.287 \\
			textureless & 2x & 0.365 & 0.648 & 0.436 & 3.717 & 4.509 & 4.069 & 0.183 & 0.235 & 0.261 \\
			textureless & 4x & 0.356 & 0.551 & 0.299 & 4.195 & 6.548 & 3.077 & 0.146 & 0.231 & 0.173 \\
			\bottomrule
		\end{tabular}
	\end{center}
\end{table*}


\begin{table*}[tb]
	\caption{Accuracy of trajectory estimates in RPE and ATE for purely event or frame-based tracking}
	\label{tab:ablation_study}
	\begin{center}
		\begin{tabular}{ccccccccccc}
			\toprule
			& &\multicolumn{3}{c}{transl. RMSE RPE in\,m} &
			\multicolumn{3}{c}{rot. RMSE RPE in\,deg} &
			\multicolumn{3}{c}{transl. RMSE ATE in\,m}\\
			\cmidrule(lr){3-5} \cmidrule(lr){6-8} \cmidrule(lr){9-11} 
			method & dataset & falling & sliding & dice & falling & sliding & dice & falling & sliding & dice \\
			\midrule
			event-based tracking & box & 0.162 & 0.046 & 0.067 & 8.182 & 2.628 & 11.172 & 0.151 & 0.024 & 0.035 \\
			event-based tracking & can & 0.045 & 0.091 & 0.039 & 5.196 & 2.745 & 3.069 & 0.021 & 0.068 & 0.027 \\
			frame-based alignment & box & 0.072 & $\infty$ & $\infty$ & 7.792 & $\infty$ & $\infty$ & 0.042 & $\infty$ & $\infty$\\
			frame-based alignment & can & $\infty$ & $\infty$ & $\infty$ & $\infty$ & $\infty$ & $\infty$ & $\infty$ & $\infty$ & $\infty$\\
			\midrule
			method & dataset & left turn & right turn & straight & left turn & right turn & straight & left turn & right turn & straight \\
			\midrule
			event-based tracking & textured 4x & 0.958 & 0.8607 & 0.833 &  15.613 & 13.590 & 4.133 & 0.701 & 0.503 & 0.785 \\
			event-based tracking & textureless 4x & 1.186 & $\infty$ & 0.978 & 11.299  & $\infty$ & 10.084 & 0.807 & $\infty$ & 0.549 \\
			frame-based alignment & textured 4x & 0.077 & 0.048 & 0.063 & 0.851 & 1.041 & 0.555 & 0.051 & 0.024 & 0.033\\
			frame-based alignment & textureless 4x & 0.149 & 0.139 & 0.062 & 2.198 & 2.609 & 0.659 & 0.066 & 0.077 & 0.036\\
			\bottomrule
		\end{tabular}
	\end{center}
\end{table*}

\subsubsection{Synthetic car sequences}

In Table~\ref{tab:accuracy_cars} we present our RPE and ATE results for the car sequences.
We use a mean SDF shape over a set of car shapes to track both cars. 
The maximum length of the shape is about 4.2\,m. 
The camera has an average distance of 8.3\,m.
Again, our approach determines position and rotation of the objects at good accuracy.
The textured object provides more events on the objects and hence can be tracked more accurately.
For the textureless object, less events are available inside the objects (see Fig.~\ref{fig:overview}) which makes the object more difficult to track. 
We see a slight degradation in accuracy in both translational and rotational motion.
Our approach handles varying speeds of the cars well with similar accuracy.
Interestingly, it is less accurate for slow cars since less events are generated and fewer event buffer frames are available for optimizing the spline segments.
We also show trajectory overlays in Fig.~\ref{fig:trajectoryoverlays} and qualitative comparisons with ground truth.

\subsection{Ablation Study}
We compare our approach with purely event- or frame-based variants.
For pure event-based tracking, we use the latest frames within the spline as keyframe. 
For pure frame-based alignment, we use all frames as keyframes and initialize the pose of new keyframes based on the previous frame. In the car and YCB sequences, there are about 900 and 2500 event buffer frames per second in average, respectively.
While frame-based alignment can outperform our method on the car sequences, it diverges for the can object early in the sequence.
For the latter, the frame differences are too large for frame-based tracking which is alleviated by the higher rate event buffer frames. 
Moreover, our event-based tracking layer provides updates on the continuous spline estimate at double the time resolution.
Our results demonstrate that the combination of both layers is advantageous for tracking. 

\subsection{Computation Time}
We measure the run-time of our approach with the textureless car sequences on an Intel Xeon Silver 4112 CPU@2.60GHz with 8 cores.
On average our implementation requires 109.5\,ms per spline segment optimization on the event-based tracking layer.
Per windowed keyframe optimization it uses 363\,ms on average.
Our implementation processes the in avg. approx. 2.4\,s trajectories in 24.5\,s.
Note that it has not been tuned yet e.g. through parallel processing.

\subsection{Real Data}
We also test our method on real data from a DAVIS240C sensor with challenging low-resolution gray scale images and noisy event measurements.
In the first frame, the tracked pose is initialized with ground truth obtained by a motion capture system.
Results on three sequences with translational (0.77\,s), rotational (0.54\,s) and circular motion (1.56\,s) can be seen in Fig.~\ref{fig:realdataresults} and in the supplementary video.
We observe that our approach requires careful tuning and accurate initialization for this noisy data.
The incremental tracking on both layers leads to drift which can make tracking fail, especially on longer sequences.
The photometric alignment error in Eq.~\eqref{eq:photometricerror} is determined in patches of size 3$\times$3 around each keypoint and a keyframe window size of $W=10$ is used.
Additionally, a velocity regularization term $E_{\mathit{reg},2} = \lambda_{\mathit{reg},2} \sum_{F \in \mathcal{F}} \Delta t_F \left\lVert \dot{\mathbf{T}}(t_F) \right\rVert_2^2$ is added with $\lambda_{\mathit{reg},2} = 1.6$.
The term favors small velocities, if the events are sparse and an event buffer frame spans larger time intervals.
We set $w_c=0.005$ for the circular motion, $w_c=0.001$ otherwise.

\section{Conclusions}

In this paper we present a novel model- and optimization-based approach for object tracking from events and frames.
We propose a two-stage processing pipeline which on the faster layer tracks the object motion from the asynchronous event stream with regard to a reference image frame.
The tracking is formulated as probabilistic inference for the object trajectory based on a generative event measurement model.
We represent the trajectory continuously using a cubic B-spline which allows us to determine object pose and velocity estimates at arbitrary times on the spline for the event measurement model.
The poses of the reference frames are estimated in a second optimization layer which optimizes the object poses in  key frames using direct image alignment on keypoints.
We evaluate our approach on synthetic sequences of typical household objects and car objects, demonstrate good tracking accuracy and analyze the combination of events and frames in an ablation study. 
We also report on experiments with real camera data.

Current  limitations  of  our  approach  include the  requirement of accurate initialization, trajectory smoothness assumption by the spline and the incremental tracking  which  leads  to  drift. If accumulated  drift or motion changes become too high, our tracking  approach  fails.

 In future work, we plan to scale our approach further for improved tracking on longer sequences and real event measurements. The robustness of our method could be improved, for instance, by combining it with object pose detection or by tight coupling of the event- and frame-based tracking.

\bibliographystyle{IEEEtran}
\bibliography{egbib}

\begin{thebibliography}{10}
\providecommand{\url}[1]{#1}
\csname url@samestyle\endcsname
\providecommand{\newblock}{\relax}
\providecommand{\bibinfo}[2]{#2}
\providecommand{\BIBentrySTDinterwordspacing}{\spaceskip=0pt\relax}
\providecommand{\BIBentryALTinterwordstretchfactor}{4}
\providecommand{\BIBentryALTinterwordspacing}{\spaceskip=\fontdimen2\font plus
\BIBentryALTinterwordstretchfactor\fontdimen3\font minus
  \fontdimen4\font\relax}
\providecommand{\BIBforeignlanguage}[2]{{%
\expandafter\ifx\csname l@#1\endcsname\relax
\typeout{** WARNING: IEEEtran.bst: No hyphenation pattern has been}%
\typeout{** loaded for the language `#1'. Using the pattern for}%
\typeout{** the default language instead.}%
\else
\language=\csname l@#1\endcsname
\fi
#2}}
\providecommand{\BIBdecl}{\relax}
\BIBdecl

\bibitem{gallego2020_evsurvey}
G.~{Gallego}, T.~{Delbruck}, G.~M. {Orchard}, C.~{Bartolozzi}, B.~{Taba},
  A.~{Censi}, S.~{Leutenegger}, A.~{Davison}, J.~{Conradt}, K.~{Daniilidis},
  and D.~{Scaramuzza}, ``Event-based vision: A survey,'' \emph{IEEE
  Transactions on Pattern Analysis and Machine Intelligence}, pp. 1--1, 2020.

\bibitem{kim2016_slamtracking}
H.~Kim, S.~Leutenegger, and A.~Davison, ``Real-time 3d reconstruction and 6-dof
  tracking with an event camera,'' in \emph{Proc. of the European Conference on
  Computer Vision (ECCV)}, 2016, pp. 349--364.

\bibitem{rebecq2017_evo}
H.~{Rebecq}, T.~{Horstschaefer}, G.~{Gallego}, and D.~{Scaramuzza}, ``Evo: A
  geometric approach to event-based 6-dof parallel tracking and mapping in real
  time,'' \emph{IEEE Robotics and Automation Letters}, vol.~2, no.~2, pp.
  593--600, 2017.

\bibitem{vasco2017_evmodet}
V.~{Vasco}, A.~{Glover}, E.~{Mueggler}, D.~{Scaramuzza}, L.~{Natale}, and
  C.~{Bartolozzi}, ``Independent motion detection with event-driven cameras,''
  in \emph{Proc. of the International Conference on Advanced Robotics (ICAR)},
  2017, pp. 530--536.

\bibitem{mitrokhin2018_evmodt}
A.~{Mitrokhin}, C.~{Ferm\"uller}, C.~{Parameshwara}, and Y.~{Aloimonos},
  ``Event-based moving object detection and tracking,'' in \emph{IEEE/RSJ
  International Conference on Intelligent Robots and Systems (IROS)}, 2019, pp.
  6105--6112.

\bibitem{mitrokhin2019_evimo}
A.~Mitrokhin, C.~F. C.~Ye, Y.~Aloimonos, and T.~Delbruck, ``{EV-IMO}: Motion
  segmentation dataset and learning pipeline for event cameras,'' in
  \emph{International Conference on Intelligent Robots and Systems (IROS)},
  2019.

\bibitem{dubeau2020_rgbde}
E.~Dubeau, M.~Garon, B.~Debaque, R.~Charette, and J.-F. Lalonde, ``{RGB-D-E}:
  Event camera calibration for fast 6-dof object tracking,'' in \emph{In
  Symposium on Mixed and Augmented Reality (ISMAR)}, 2020, pp. 127--135.

\bibitem{bryner2019_eventbasedtracking}
S.~{Bryner}, G.~{Gallego}, H.~{Rebecq}, and D.~{Scaramuzza}, ``Event-based,
  direct camera tracking from a photometric 3d map using nonlinear
  optimization,'' in \emph{2019 International Conference on Robotics and
  Automation (ICRA)}, 2019, pp. 325--331.

\bibitem{corke2013_rvc}
P.~Corke, \emph{Robotics, Vision and Control: Fundamental Algorithms in
  MATLAB}, 1st~ed.\hskip 1em plus 0.5em minus 0.4em\relax Springer Publishing
  Company, Incorporated, 2013.

\bibitem{patronperez2015_splinetraj}
A.~Patron-Perez, S.~Lovegrove, and G.~Sibley, ``A spline-based trajectory
  representation for sensor fusion and rolling shutter cameras,''
  \emph{International Journal of Computer Vision}, vol. 113, no.~3, pp.
  208--219, Jul. 2015.

\bibitem{engel2018_dso}
J.~{Engel}, V.~{Koltun}, and D.~{Cremers}, ``Direct sparse odometry,''
  \emph{IEEE Transactions on Pattern Analysis and Machine Intelligence},
  vol.~40, no.~3, pp. 611--625, 2018.

\bibitem{wang2020_directshape}
R.~Wang, N.~Yang, J.~St\"uckler, and D.~Cremers, ``{DirectShape}: Photometric
  alignment of shape priors for visual vehicle pose and shape estimation,'' in
  \emph{Proc. of the IEEE International Conference on Robotics and Automation
  (ICRA)}, 2017.

\bibitem{rublee2011_orb}
E.~{Rublee}, V.~{Rabaud}, K.~{Konolige}, and G.~{Bradski}, ``Orb: An efficient
  alternative to sift or surf,'' in \emph{International Conference on Computer
  Vision (ICCV)}, 2011, pp. 2564--2571.

\bibitem{prisacariu2012_segposerecon}
V.~A. Prisacariu, A.~V. Segal, and I.~Reid, ``Simultaneous monocular {2D}
  segmentation, {3D} pose recovery and {3D} reconstruction,'' in \emph{Proc. of
  the Asian Conf. on Computer Vision (ACCV)}, 2013.

\bibitem{dame2013_denserecon}
A.~Dame, V.~A. Prisacariu, C.~Y. Ren, and I.~D. Reid, ``Dense reconstruction
  using {3D} object shape priors,'' in \emph{Proc. of the IEEE Int. Conf. on
  Computer Vision and Pattern Recognition (CVPR)}, 2013, pp. 1288--1295.

\bibitem{leutenegger2015_okvis}
S.~Leutenegger, S.~Lynen, M.~Bosse, R.~Siegwart, and P.~Furgale,
  ``Keyframe-based visual–inertial odometry using nonlinear optimization,''
  \emph{International Journal of Robotics Research}, vol.~34, no.~3, pp.
  314--334, Mar. 2015.

\bibitem{sturm2012_iros}
J.~Sturm, N.~Engelhard, F.~Endres, W.~Burgard, and D.~Cremers, ``A benchmark
  for the evaluation of {RGB-D SLAM} systems,'' in \emph{Proc. of the
  International Conference on Intelligent Robot Systems (IROS)}, 2012, pp.
  573--580.

\bibitem{rebecq2018_cvsimulator}
H.~Rebecq, D.~Gehrig, and D.~Scaramuzza, ``{ESIM}: an open event camera
  simulator,'' \emph{Proc. of the 2nd Conf. on Robotics Learning (CoRL)}, pp.
  969--982, 2018.

\bibitem{calli2015_ycb}
B.~{Calli}, A.~{Singh}, A.~{Walsman}, S.~{Srinivasa}, P.~{Abbeel}, and A.~M.
  {Dollar}, ``The ycb object and model set: Towards common benchmarks for
  manipulation research,'' in \emph{International Conference on Advanced
  Robotics (ICAR)}, 2015, pp. 510--517.

\bibitem{chang2015_shapenet}
\BIBentryALTinterwordspacing
A.~X. Chang, T.~Funkhouser, L.~Guibas, P.~Hanrahan, Q.~Huang, Z.~Li,
  S.~Savarese, M.~Savva, S.~Song, H.~Su, J.~Xiao, L.~Yi, and F.~Yu,
  ``{ShapeNet}: An information-rich 3d model repository,'' 2015. [Online].
  Available: \url{http://arxiv.org/abs/1512.03012}
\BIBentrySTDinterwordspacing

\end{thebibliography}

\end{document}